\documentclass[conference]{IEEEtran}
\IEEEoverridecommandlockouts
\usepackage{cite}
\usepackage{enumitem}  
\usepackage{float} 
\usepackage{amsmath,amssymb,amsfonts}
\usepackage{algorithmic}
\usepackage{graphicx}
\usepackage{textcomp}
\usepackage{xcolor}
\usepackage{mathptmx}
\usepackage{amsmath}
\usepackage{graphicx}
\usepackage{booktabs}

\usepackage[T1]{fontenc} 
\def\BibTeX{{\rm B\kern-.05em{\sc i\kern-.025em b}\kern-.08em
    T\kern-.1667em\lower.7ex\hbox{E}\kern-.125emX}}
\begin{document}

\title{Learning Dynamic Scene Reconstruction with Sinusoidal Geometric Priors}

\author{
\IEEEauthorblockN{Tian Guo\textsuperscript{1}, Hui Yuan\textsuperscript{1}, Philip Xu\textsuperscript{1}, and David Elizondo\textsuperscript{1}}
\IEEEauthorblockA{
\textsuperscript{1}De Montfort University, Leicester, United Kingdom\\
Email: \{tian.guo, hui.yuan, philip.xu, david.elizondo\}@dmu.ac.uk
}
\thanks{Corresponding author: rhamzaoui@dmu.ac.uk}
}

\maketitle

\begin{abstract}
This paper proposes a novel loss function "SirenPose" that innovatively combines the periodic activation characteristics of SIREN (Sinusoidal Representation Networks) with geometric prior information of keypoint structures to improve the accuracy of dynamic 3D scene reconstruction. Existing methods often struggle to maintain motion modeling accuracy and spatiotemporal consistency when handling fast-moving multi-target scenes. By introducing physics-based constraint mechanisms, SirenPose ensures coherence of keypoint predictions in both spatial and temporal dimensions. We expanded the training dataset to 600,000 annotated instances. Experimental results show that models based on SirenPose achieved significant improvements in spatiotemporal consistency metrics compared to existing methods, demonstrating superior performance in handling rapid motion and complex scene changes. 
\end{abstract}

\section{Introduction}
In computer vision, reconstructing dynamic 3D scenes from video sequences remains a fundamental and crucial task\cite{3Dgs,z1,z2,chongjian5,优化器,z22}. Traditional computer graphics approaches primarily rely on multi-view\cite{chongjian2,Z3,z4,chongjian4,z17} reconstruction and template-based techniques, which require specialized hardware setups or complex multi-camera arrays, thus limiting their practical applications. With the advancement of deep learning technologies, learning-based dynamic scene reconstruction methods have demonstrated unique advantages in reconstructing 3D structures and capturing complex dynamic changes, such as human motion and object deformation, from monocular video sequences. These technologies have revolutionized creative industries, showing tremendous potential in film visual effects, game development, and virtual reality applications\cite{VR,Z5,z7}.
Recent breakthroughs in video-to-3D generation methods\cite{nerf}\cite{4DGS}\cite{chongjian4} based on predictive generative models have significantly advanced dynamic content creation. Current approaches predominantly rely on multi-view data for scene reconstruction, combining differentiable rendering techniques such as Neural Radiance Fields (NeRF)\cite{nerf} or Gaussian Splatting with generative models based on Score Distillation Sampling (SDS)\cite{SDS}\cite{clip}. While recent studies have begun exploring monocular video-based 3D dynamic\cite{4DGS} scene reconstruction, incorporating SE(3)\cite{Shape-Motion} motion basis representations for explicit modeling of long-term 3D motion trajectories and employing "decomposition-reconstruction"\cite{DreamScene4D,z6} strategies for complex scenes with multiple dynamic objects, significant challenges remain in handling real-world dynamic scenes. Particularly when scenes contain rapidly moving multiple objects or frequent occlusions and interactions, existing methods show limitations in two critical aspects: motion modeling accuracy and spatiotemporal consistency. In terms of motion modeling precision, current approaches primarily rely on discrete pose keypoints or SE(3)-based motion basis functions, struggling to capture detailed features during rapid movements and non-rigid deformations, resulting in trajectory jitter and temporal inconsistencies. Additionally, the lack of effective geometric constraints makes it difficult to maintain 3D structural consistency during object occlusions and multi-object\cite{Nerf++} interactions, often leading to penetration and incorrect occlusion issues\cite{zhedang}. These limitations significantly impact the practical effectiveness in high-precision applications such as virtual reality.
latex
\begin{figure}[!t]
\centering
\includegraphics[width=\columnwidth]{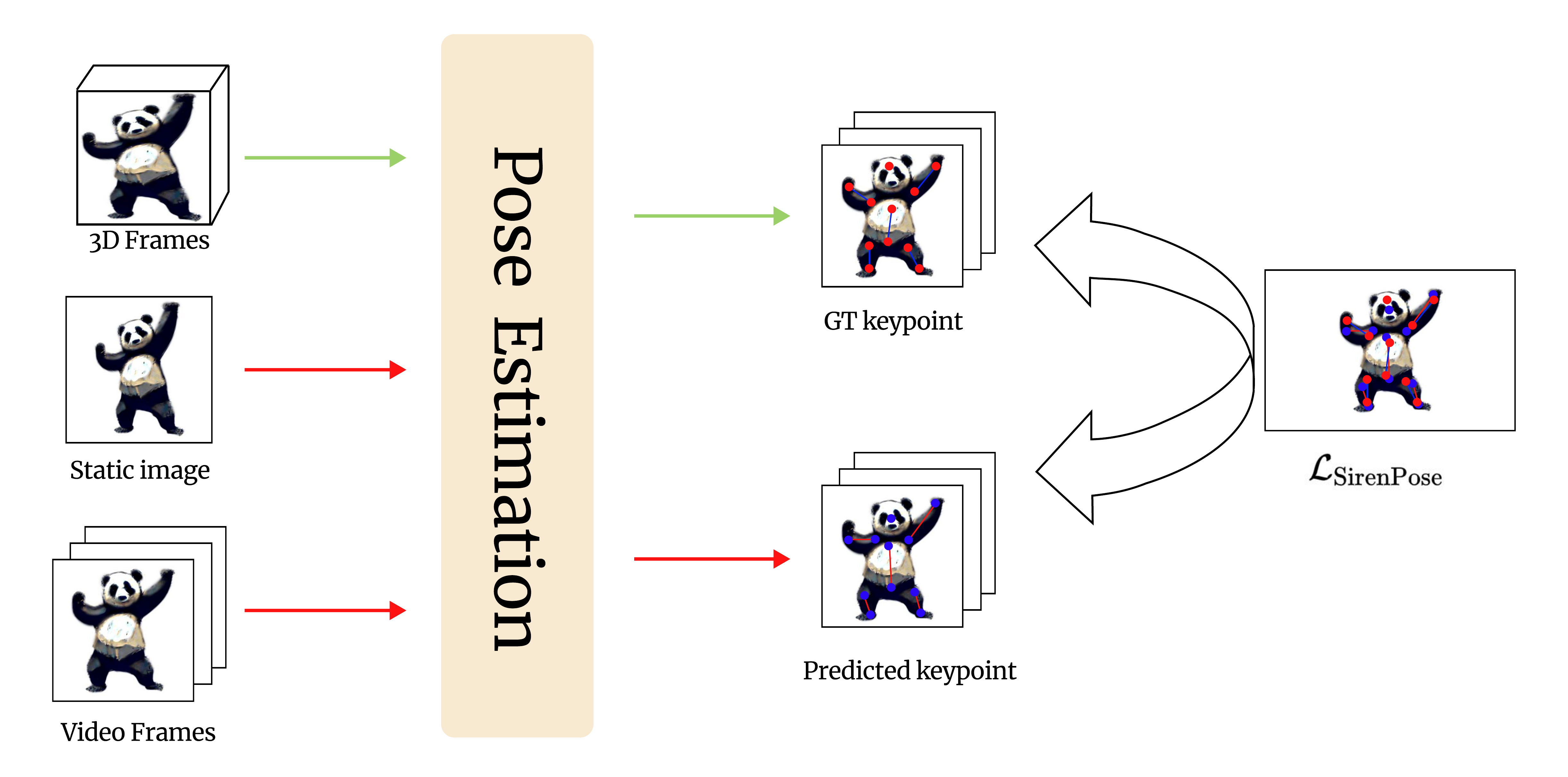}
\caption{This figure illustrates the overall workflow of the SirenPose framework. At the input stage, the system accepts three different types of data: 3D frames, static images, and video frames. All input data are processed by the “Pose Estimation” module to extract the target's keypoint information. Subsequently, the predicted keypoints are compared with the ground truth keypoints (GT Keypoint). Using the loss function $L_{\text{SirenPose}}$, the framework optimizes the keypoint predictions to ensure accuracy and consistency. This process adapts to various input formats while achieving efficient pose estimation.}
\label{fig:workflow}
\end{figure}
In this paper, we present "SirenPose," a novel loss function that leverages the periodic activation characteristics of SIREN\cite{SIREN1}\cite{SIREN123} (Sinusoidal Representation Networks) and the geometric priors of keypoint\cite{x-pose,z8} structural correlations - inherent geometric constraints exist between object keypoints, and this structured information can be more accurately expressed through SIREN\cite{SIRENfanhua} networks' high-frequency modeling capabilities. By enforcing this structural consistency, we introduce a powerful, physics-based constraint mechanism that ensures predicted keypoints maintain coherence in both spatial and temporal dimensions. This approach enables precise capture of detailed features during rapid motion while preserving geometric constraints between keypoints, effectively addressing complex scenarios involving occlusions and fast movements. This methodology has enabled stable training on large-scale\cite{zhedang2}\cite{zhedang3} dynamic scene datasets, including various motion scenarios, multi-view capture data, and in-the-wild videos with complex backgrounds. Our method demonstrates significant improvements in dynamic scene reconstruction accuracy, reducing motion trajectory jitter and discontinuities while maintaining natural scene transitions\cite{chongjianyewai}.

Our approach enables the integration of the UniKPT dataset from X-Pose\cite{x-pose} (338 keypoints, 1,237 categories, 400K instances) with additionally collected data, expanding the training set to 600K annotated instances. This enriched dataset provides more reliable supervision signals for dynamic scene reconstruction\cite{guangdingsam}. Experimental results demonstrate that SirenPose significantly enhances the model's ability to capture high-frequency features and maintain spatiotemporal consistency in dynamic scenes. In standard video reconstruction benchmarks, SirenPose-based methods show approximately 6\% improvement in spatiotemporal consistency metrics compared to existing approaches, demonstrating superior robustness in handling rapid motion and complex scene changes. This validates the innovation and effectiveness of SirenPose as a core loss design for dynamic scene reconstruction.

Our key contributions are summarized as follows:
\begin{enumerate}
    \item We propose the novel SirenPose loss function, combining SIREN networks' periodic activation characteristics with geometric priors of keypoint structures to effectively supervise video reconstruction of dynamic scenes.
    \item We establish a large-scale keypoint supervision training framework, integrating and expanding to 600K annotated instances, utilizing graph neural networks to model keypoint constraint relationships.
    \item We introduce a high-frequency feature supervision strategy that significantly improves detail fidelity and spatiotemporal consistency\cite{SIRENfanhua} in dynamic scene reconstruction through the SirenPose loss function.
    \end{enumerate} 
\section{Related Work}
\subsection{Video-to-3D Reconstruction}
Dynamic scene reconstruction has evolved from multi-view to monocular video approaches\cite{chongjian1}\cite{chongjian2}\cite{chongjian3}\cite{chongjian4}\cite{chongjian5}. Early methods primarily relied on multi-view videos and precise camera calibration, achieving reconstruction of visible areas through techniques like Dynamic NeRF \cite{nerf}or Dynamic Gaussian Splatting\cite{3Dgs}. These methods performed well in static or slow-moving scenes but required complex capture equipment and calibration processes.Recent research has shifted towards the more challenging task of monocular video reconstruction. The key breakthrough lies in decomposing complex scene dynamics into more manageable components: camera motion, object deformation, and rigid transformations. This decomposition strategy\cite{CoTracker} not only simplifies the optimization process but also allows separate handling of different motion patterns\cite{Shape-Motion,z11}. By incorporating physical priors and data-driven supervision signals, new methods can better handle complex scenarios such as rapid motion, multi-object scenes, and occlusions, while achieving reasonable completion of unobserved regions.
\subsection{Dynamic Keypoint Representation}
Keypoint detection and motion modeling are fundamental for understanding dynamic scenes\cite{unipose,x-pose,z9,z10}. Traditional methods are often limited to specific categories (such as humans or animals) and struggle to generalize to new categories. Recent work has introduced the category-agnostic keypoint detection (CAPE) \cite{CAPE}paradigm, which requires only a small number of annotated supporting images to achieve keypoint localization for any category.
Recent advancements indicate that modeling keypoints as a graph structure rather than as independent entities offers significant advantages. By utilizing graph neural networks, we can effectively leverage the geometric constraints and structural correlations among keypoints, which not only helps break symmetry but also better handles complex situations like occlusion. Additionally, combining multimodal cues\cite{SAM2} (such as visual and textual information) can significantly enhance the accuracy and generalization of keypoint localization.
\subsection{SIREN in High-Frequency Neural Representation}
SIREN (Sinusoidal Representation Networks) employs periodic activation functions\cite{SIREN1,SIREN12}, demonstrating unique advantages in modeling high-frequency signals. Compared to traditional activation functions like ReLU, SIREN can more accurately represent geometric shapes and continuous signal features\cite{SIREN123}.
Research shows that the periodic characteristics of SIREN provide a natural advantage in capturing complex spatial relationships and geometric constraints. By incorporating structural prior knowledge of keypoints, it can more effectively guide the expression of high-frequency details in dynamic scene reconstruction, opening new avenues for enhancing spatiotemporal consistency and improving motion modeling accuracy.
\section{Approach}
Our objective is to enhance the accuracy and spatio-temporal consistency of dynamic scene reconstruction by introducing the "SirenPose" loss function. This method innovatively combines the periodic activation characteristics of SIREN (Sinusoidal Representation Networks)\cite{SIREN1} with geometric prior information of keypoints to optimize the reconstruction quality of dynamic 3D scenes. Input: A set of predicted keypoint positions $\hat{K} = {\hat{k}i}{i=1}^M$ in dynamic scenes and the corresponding ground truth keypoint positions $K = {k_i}_{i=1}^M$, where $M$ represents the total number of keypoints. Output: Optimized keypoint position sequences for achieving high-precision dynamic scene reconstruction. To achieve the above objectives, this research proposes the following key technical components: A.Model Theoretical Foundation. B.Parameter Initialization and Training Strategy. C.SIREN’s Periodic Activation and Keypoint Geo-
metric Priors. D.SirenPose Loss Function. E.Supervision Mechanism and Backpropagation.

\subsection{Model Theoretical Foundation}
Dynamic scene reconstruction aims to reconstruct spatiotemporally continuous 3D scenes, which consist of low-frequency components describing global structures and high-frequency components capturing details and rapid changes. Traditional methods, such as CAPE, focus on modeling low-frequency components, often neglecting high-frequency details. SIREN (Sinusoidal Representation Networks) addresses this by effectively capturing high-frequency signals. The optimization objective, combining CAPE and SirenPose models, is expressed as:
\begin{multline}
f_{\text{CAPE-SirenPose}}(x,t) = f_{\text{CAPE}}(x,t) + \lambda f_{\text{SIREN}}(x,t) \\
\approx f_{\text{low}}(x,t) + \lambda f_{\text{high}}(x,t),
\end{multline}
where $\lambda$ controls the importance of high-frequency components. This allows the model to capture both global structures and fine details, ensuring accurate dynamic scene reconstruction.SIREN's sinusoidal activation function, $\sigma(z) = \sin(\omega_0 z)$, efficiently models high-frequency signals, outperforming traditional activation functions like ReLU. When combined with geometric priors of keypoints, it further enhances detail capture and reconstruction accuracy. Keypoints in dynamic scenes (e.g., joints or features) have geometric constraints, such as lengths and angles, expressed as $d_{ij} = \lvert k_i - k_j \rvert$. Incorporating these priors ensures structural consistency, helping the model handle missing data or noise and accurately reconstruct complex dynamic changes.
\subsection{Parameter Initialization and Training Strategy}
To fully exploit SIREN's ability\cite{SIREN12,z12} to represent high-frequency details, specific parameter initialization methods are applied. For the first layer, weights are initialized as:
\[
W^0 \sim \mathcal{U}\left(-\dfrac{1}{\omega_0}, \dfrac{1}{\omega_0}\right),
\]
where $\omega_0 = 30$ ensures rich periodicity of the sine function. For subsequent layers ($l > 0$), weights are initialized as $W^l \sim \mathcal{U}(-\sqrt{6/n_l}, \sqrt{6/n_l})$, where $n_l$ denotes the input dimension of the $l$-th layer. This initialization ensures stable training and effective representation of both low- and high-frequency components.

The training process directly incorporates the combined loss function, which includes the reconstruction loss $\mathcal{L}_{\text{recon}}$ and the SirenPose loss weighted by $\lambda_{\text{sp}}$. These losses are optimized simultaneously during training using adaptive optimizers such as Adam. This approach ensures that the model balances global structure reconstruction with high-frequency detail consistency, achieving accurate and spatiotemporally coherent dynamic scene reconstruction.
\subsection{SIREN's Periodic Activation and Keypoint Geometric Priors}

SIREN employs sinusoidal functions as activation functions, defined as $\sigma(z) = \sin(\omega_0 z)$, where $\omega_0$ is the frequency factor controlling the oscillation frequency. The hierarchical representation of SIREN network processes the input $x$ through L layers, with each layer applying a sinusoidal transformation:
\[
h^0 = x, \quad h^l = \sin\left(\omega_0 W^l h^{l-1} + b^l\right), \quad l = 1, ..., L,
\]
\begin{figure*}
    \includegraphics[width=\textwidth, height=0.36\textheight]{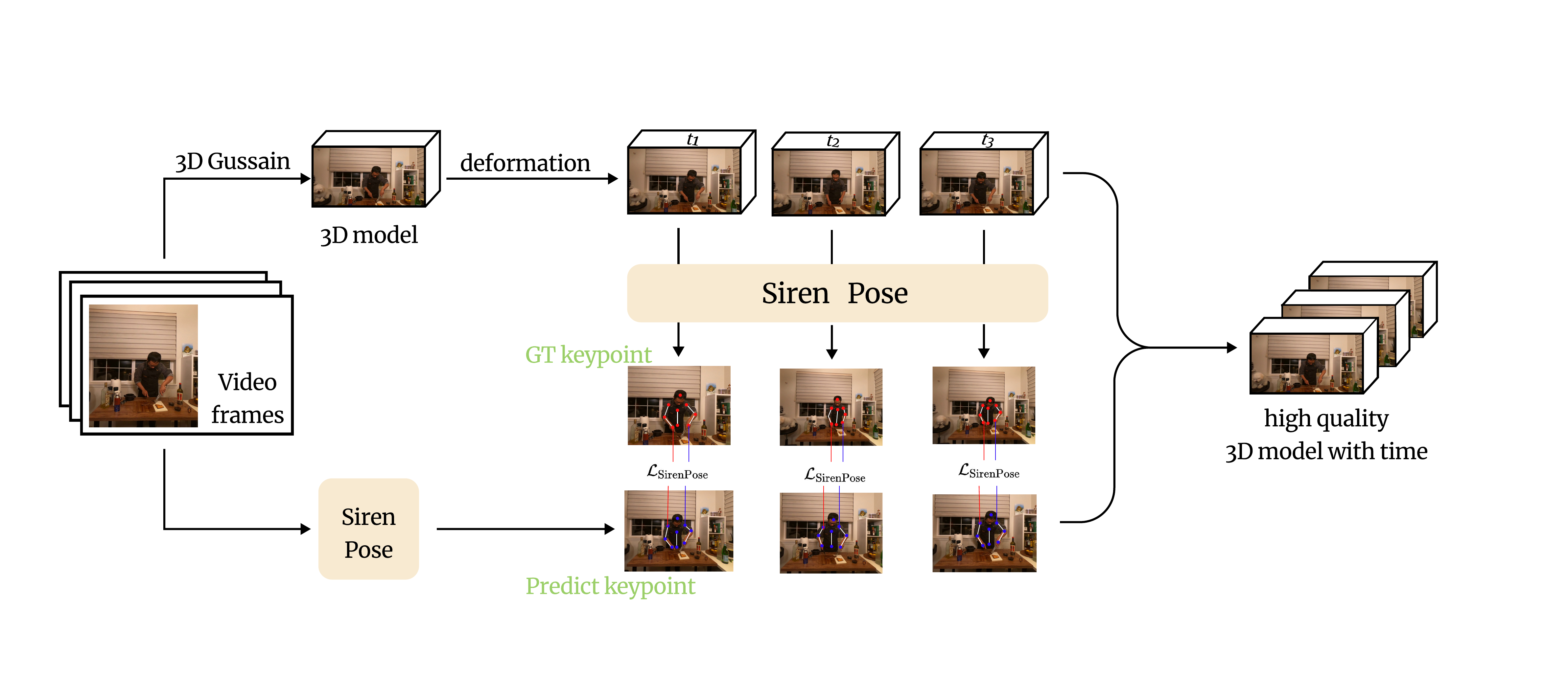}
    \caption{As shown in the figure, our proposed SirenPose architecture consists of the following innovative components: (1) Input video frames are first processed through 3D Gaussian to obtain initial 3D model representations, followed by deformation processing to generate temporally-related 3D model sequences; (2) The core Siren Pose module innovatively combines SIREN's periodic activation characteristics with geometric prior information of keypoints, while simultaneously extracting GT keypoints and predicted keypoints from the input video; (3) Through a carefully designed $L_{SirenPose}$ loss function, this architecture can effectively optimize both spatial localization accuracy and temporal consistency of keypoints, ultimately outputting high-quality and spatio-temporally continuous dynamic 3D models. Experiments demonstrate that this end-to-end architectural design shows significant advantages in handling rapid motion and complex scene changes.}
    \label{fig:pipeline}
\end{figure*}
where $W^l$ and $b^l$ are learnable weights and biases at layer $l$, and $h^l$ represents the feature embedding at each layer. This periodic activation enables effective modeling of complex signals across different frequency bands.To maintain structural consistency, we incorporate geometric constraints between keypoints\cite{fenceng1} in dynamic scenes. For any pair of keypoints $k_i$ and $k_j$, their spatial relationship is captured through the distance\cite{D-Nerf} metric:
\[
d_{ij} = \|k_i - k_j\| = \sqrt{\sum_{d=1}^D (k_i^{(d)} - k_j^{(d)})^2} = \lvert k_i - k_j \rvert,
\]

where $d_{ij}$ measures the Euclidean distance between keypoints across D dimensions. These geometric relationships serve as prior knowledge in the loss function\cite{ziji}, helping preserve natural motion constraints like bone lengths and joint angles during scene reconstruction.
\subsection{SirenPose Loss Function}
The SirenPose loss integrates two objectives: position accuracy and geometric consistency\cite{yueshu}. The position term minimizes Euclidean distances between predicted keypoints $\hat{k}_i$ and ground truth keypoints $k_i$ to ensure accurate localization. The geometric term, weighted by $\lambda_{\text{geo}}$, enforces structural relationships between keypoint pairs $E$ through sinusoidal functions with frequency $\omega_0$. This combination enables precise keypoint positioning while maintaining consistent geometric relationships:

\begin{multline}
\mathcal{L}_{\text{SirenPose}} = \sum_{i=1}^M \|\hat{k}_i - k_i\|_2^2 \\
+ \lambda_{\text{geo}} \sum_{(i,j) \in E} 
\Big\| \sin\big(\omega_0 (\hat{k}_i - \hat{k}_j)\big) 
- \sin\big(\omega_0 (k_i - k_j)\big) \Big\|_2^2
\end{multline}
\subsection{Supervision Mechanism and Backpropagation}
The model training is guided by a composite loss function that combines reconstruction loss and SirenPose loss:
\[
\mathcal{L} = \mathcal{L}_{\text{recon}} + \lambda_{\text{sp}} \mathcal{L}_{\text{SirenPose}}
\]
where $\mathcal{L}_{\text{recon}} = \left\| f_{\text{pred}}(x, t) - f_{\text{target}}(x, t) \right\|_2^2$ measures the difference between predicted and target scenes, and $\lambda_{\text{sp}}$ balances the contribution of SirenPose loss. During backpropagation, gradients are computed with respect to model parameters $\theta$, incorporating both reconstruction and geometric consistency objectives:
\[
\frac{\partial \mathcal{L}}{\partial \theta} = \frac{\partial \mathcal{L}_{\text{recon}}}{\partial \theta} + \lambda_{\text{sp}} \frac{\partial \mathcal{L}_{\text{SirenPose}}}{\partial \theta}
\]
For each predicted keypoint $\hat{k}_i$, the gradient combines positional error and geometric consistency terms, driving predictions toward ground truth positions while maintaining structural relationships between keypoints.

\section{ Experiment}
To comprehensively validate the effectiveness of SirenPose, we conducted extensive experiments on multiple public datasets, including the DAVIS dataset which is widely used for dynamic scene evaluation. DAVIS (Densely Annotated VIdeo Segmentation) is a high-quality video dataset that focuses on object segmentation and motion analysis in dynamic scenes, covering complex scenarios such as rapid motion, multi-object interactions, and occlusions. Due to its fine-grained annotations and diverse scene types, DAVIS has been widely adopted for evaluating video analysis, optical flow estimation, and dynamic scene reconstruction tasks\cite{z13,z14,SIRENfanhua,z15}.
All experiments were conducted on a single NVIDIA A6000 GPU (48GB). To ensure fair experimental comparisons, we strictly followed the configuration parameters of DreamScene4D. During model training, we employed the Adam optimizer with an initial learning rate of $1\times10^{-4}$ and a batch size of 64.
\subsection{Video to Reconstruction}
To evaluate our model's capability in reconstructing 3D geometry from video inputs, we conducted comparative experiments against recent state-of-the-art methods including MoSCA, DSGS, and 4DGS. The experiments were performed on dynamic video sequences from the DAVIS dataset, focusing on both reconstruction quality and temporal consistency.
\begin{figure*}
    \includegraphics[width=\textwidth, height=0.25\textheight]{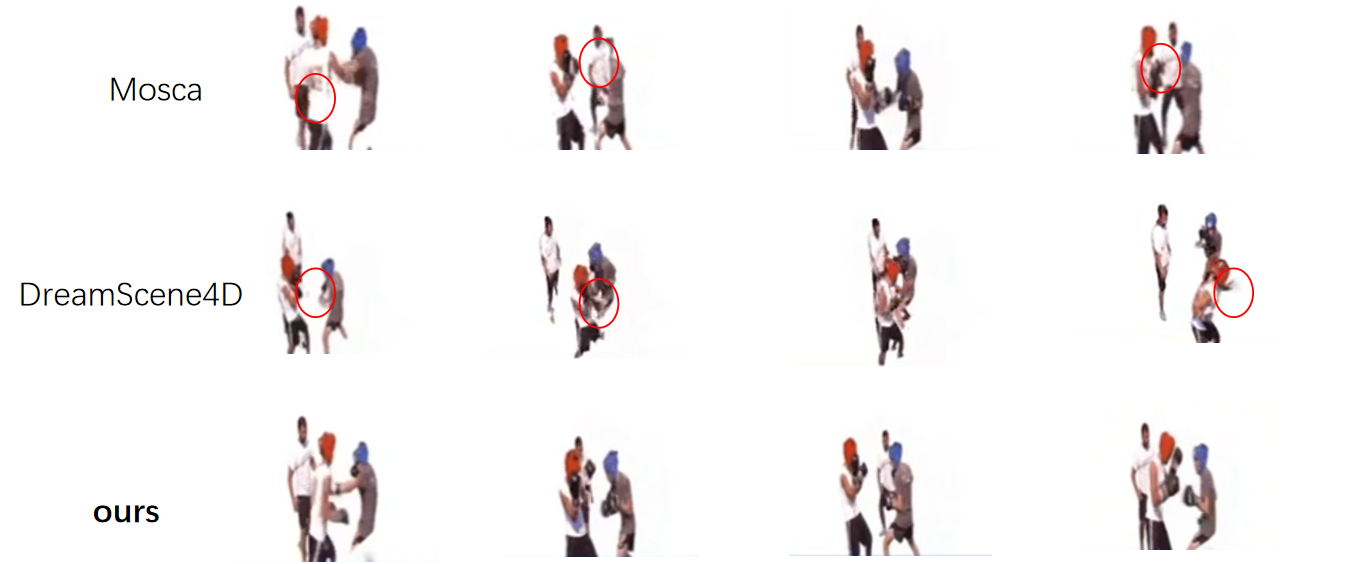}
    \caption{The figure presents qualitative comparisons between our proposed SirenPose method and two existing approaches, Mosca and DreamScene4D, on a challenging dynamic scene reconstruction task. Each row shows reconstruction results from the respective methods across multiple video frames, demonstrating their ability to handle complex motion and deformation.}
    \label{fig:pipeline}
\end{figure*}

\begin{table}[ht]
\centering
\begin{tabular}{lcc}
\toprule
\textbf{Model} & \textbf{MSE-Score} & \textbf{EPE-Score} \\
\midrule
PIPS++        & 76.0 & 83.1 \\
CoTracker     & 87.6 & 86.2 \\
DreamScene4D  & 80.2 & 84.6 \\
Ours          & \textbf{93.7*} & \textbf{90.7*} \\
\bottomrule
\end{tabular}
\caption{Quantitative comparison on MSE and EPE metrics. * indicates the best performance.}
\label{tab:epe_mse}
\end{table}

\subsection{3D Gaussian Movement Accuracy}
\begin{figure*}[!t]
    \centering
    \includegraphics[width=\textwidth, height=0.27 \textheight]{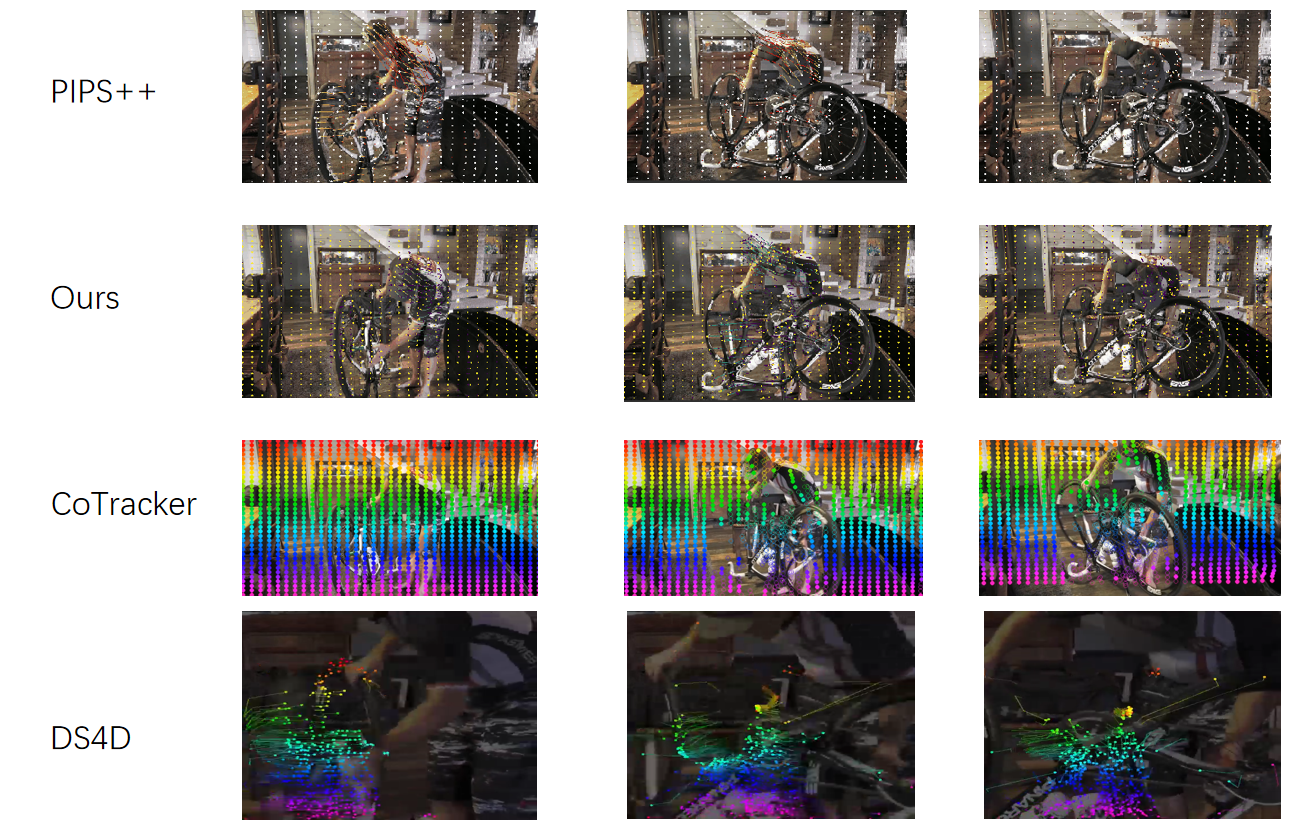}
    \caption{This figure presents a qualitative comparison of pixel-level tracking capabilities across three methods in challenging dynamic scenes. The visualization tracks specific points (indicated by red circles) through multiple frames, demonstrating each method's ability to maintain temporal consistency during complex motions and interactions.}
    \label{fig:pipeline}
\end{figure*}
\begin{table}[ht]
\centering
\caption{Comparison of Dynamic Scene Reconstruction Methods.}
\label{tab:comparison_reconstruction}
\resizebox{\columnwidth}{!}{%
\begin{tabular}{lccc}
\toprule
\textbf{Method}         & \textbf{Temporal Consistency↑} & \textbf{Geometric Accuracy↑} & \textbf{User-Score↑} \\ 
\midrule
Mosca                   & 0.72                           & 0.68                          & 69.4                \\ 
DreamScene4D            & 0.81                           & 0.75                          & 74.1                \\ 
Ours (SirenPose)        & \textbf{0.91}                  & \textbf{0.87}                 & \textbf{80.3}       \\ 
\bottomrule
\end{tabular}%
}
\end{table}

To evaluate the motion information perception and fitting performance of SirenPose on 2D videos and 3D sequences with temporal information, we conducted Optical Flow Estimation experiments on the Davis dataset. The experiments compared models including PIPS++\cite{pipdata++}, Cotracking, and DSGS, using Endpoint Error (EPE) as a quantitative metric.  Additionally, we analyzed the performance differences among these models through detailed qualitative visualizations of specific optical flow points\cite{GUANGLIU}.
\begin{table}[h]
\centering
\caption{Ablation Study on the Impact of Removing SirenPose. The evaluation metrics include PSNR (Peak Signal-to-Noise Ratio), SSIM (Structural Similarity Index), and LPIPS (Learned Perceptual Image Patch Similarity).}
\label{tab:ablation}
\begin{tabular}{lccc}
\hline
\textbf{Method}          & \textbf{PSNR↑} & \textbf{SSIM↑} & \textbf{LPIPS↓} \\ \hline
Full Model (With SirenPose) & 32.45         & 0.921         & 0.082          \\ 
Without SirenPose          & 30.12         & 0.894         & 0.115          \\ \hline
\end{tabular}
\end{table}
This experiment evaluates SirenPose's advantages in motion processing and generalization across tasks. Unlike DSGS, which relies on mask-based tracking \cite{yanmal, SAM2} and is limited to specific objects \cite{shixuyizhi}, SirenPose leverages temporal consistency for superior adaptability and motion fitting, achieving performance comparable to CoTracker. These results demonstrate its potential for dynamic scene understanding.

\subsection{Ablation study}
To evaluate the contribution of SirenPose to the overall model performance, we conducted an ablation study by removing this module and comparing the results against the full model. The metrics used for evaluation include PSNR (Peak Signal-to-Noise Ratio), SSIM (Structural Similarity Index), and LPIPS (Learned Perceptual Image Patch Similarity).

\section{Summary}
This paper introduces SirenPose, a novel framework for dynamic scene reconstruction that enhances temporal consistency and geometric accuracy. By integrating a robust pose estimation module with the specialized $L_{\text{SirenPose}}$ loss function, our method achieves precise motion representation and reconstruction across various scenarios. Comprehensive experiments demonstrate that SirenPose outperforms existing methods in handling complex interactions, with significant improvements in temporal coherence and geometric accuracy. The framework's ability to generalize across dynamic tasks highlights its potential for real-world applications in augmented reality, robotics, and motion analysis.

\bibliographystyle{IEEEbib}
\bibliography{icme2025references}

\vspace{12pt}
\color{red}
\end{document}